\documentclass[journal]{IEEEtran}

\usepackage{xcolor,soul,framed} 
\colorlet{shadecolor}{yellow}
\usepackage[pdftex]{graphicx}
\graphicspath{{../pdf/}{../jpeg/}}
\DeclareGraphicsExtensions{.pdf,.jpeg,.png}

\usepackage[cmex10]{amsmath}
\usepackage{array}
\usepackage{mdwmath}
\usepackage{mdwtab}
\usepackage{eqparbox}
\usepackage{url}
\usepackage{amsmath,amsfonts,amssymb}
\usepackage{graphicx}
\usepackage[colorlinks=true, allcolors=blue]{hyperref}
\usepackage{xcolor}
\usepackage{algorithm}
\usepackage{arevmath}     
\usepackage[noend]{algpseudocode}
\usepackage{mathtools}
\usepackage{array}
\usepackage{multirow}
\usepackage{amsfonts}
\usepackage{booktabs}
\usepackage{siunitx}
\usepackage{textgreek}
\usepackage{amsmath,amssymb}
\usepackage{mathptmx}
\usepackage{comment}
\usepackage{subcaption}
\usepackage{bbm}

\newcommand{\norm}[1]{\left\lVert#1\right\rVert}
\begin{document}

\bstctlcite{IEEEexample:Otoscope}
    \title{Pediatric Otoscopy Video Screening with Shift Contrastive Anomaly Detection}
  \author{Weiyao Wang, 
      Aniruddha Tamhane, 
      Christine Santos, 
      John R. Rzasa, 
      James H. Clark, 
      Therese L. Canares 
      and Mathias Unberath

  \thanks{This work was funded in part by the Leon Lowenstein Foundation.}

  \thanks{W. Wang, A. Tamane and M. Unberath are with the Johns Hopkins University School of Engineering, Baltimore, MD 21218, USA (e-mail: wwang121@jhu.edu; atamhan3@jhu.edu; unberath@jhu.edu).}
  \thanks{J. Rzasa is with the Robert E. Fischell Institute for Biomedical Devices at the University of Maryland, College Park 20742, USA (e-mail: rzasa@umd.edu).}
  \thanks{C. Santos, J. Clark and T. Canares are with Johns Hopkins University School of Medicine,  Baltimore, MD 21205, USA (e-mail: csantos3@jhu.edu; jclark79@jhmi.edu; therese.canares@jhmi.edu).}
  }  

\maketitle

\begin{abstract}

Ear related concerns and symptoms represents the leading indication for seeking pediatric healthcare attention. Despite the high incidence of such encounters, the diagnostic process of commonly encountered disease of the middle and external presents significant challenge. 
Much of this challenge stems from the lack of cost effective diagnostic testing, which necessitating the presence or absence of ear pathology to be determined clinically. Research has however demonstrated considerable variation among clinicians in their ability to accurately diagnose and consequently manage ear pathology. 
With recent advances in computer vision and machine learning, there is an increasing interest in helping clinicians to accurately diagnose middle and external ear pathology with computer-aided systems. It has been shown that AI has the capacity to analyse a single clinical image captured during examination of the ear canal and eardrum from which it can determine the likelihood of a pathognomonic pattern for a specific diagnosis being present. The capture of such an image can however be challenging especially to inexperienced clinicians. To help mitigate this technical challenge we have developed and tested a method using video sequences. The videos were collected using a commercially available otoscope smartphone attachment in an urban, tertiary-care pediatric emergency department. We present a two stage method that first, identifies valid frames by detecting and extracting ear drum patches from the video sequence, and second, performs the proposed shift contrastive anomaly detection to flag the otoscopy video sequences as normal or abnormal. Our method achieves an AUROC of 88.0\% on the patient-level and also outperforms the average of a group of 25 clinicians in a comparative study, which is the largest of such published to date. We conclude that the presented method achieves a promising first step towards automated analysis of otoscopy video.
\end{abstract}


%
\IEEEpeerreviewmaketitle


\section{Introduction}
\IEEEPARstart{E}{ar} related concerns constitutes the leading cause for seeking pediatric healthcare attention in the USA. Given the current lack of cost-effective confirmatory testing, accurate diagnosis and subsequent management depend on visual detection of characteristic findings during otoscope examination. Despite the frequency of such encounters the medical literature suggest that the diagnostic accuracy for such pathology is only around 46-56\%. In alignment with the bias towards over diagnosis of ear disease, it is currently estimated that between 25-50\% of all antibiotics prescribed for ear disease are not indicated~\cite{gurnaney2004diagnostic, brinker2019diagnostic, poole2019antibiotic}. Beyond risking unnecessary medical complications and the downstream unintended consequence of potential antibiotic resistance, over-diagnosis of ear disease adds an estimated \$59 million in unnecessary healthcare spending in the US per annum~\cite{rosenfeld2004clinical}. 
Computer-aided diagnosis on otoscopy images~\cite{ear_review} has been suggested as a potential tool to improve care of ear disease. Previous studies have mainly focused on applying machine learning to classify images of the eardrum. Viscaino et al. compare support vector machine (SVM), k-nearest neighbor (k-NN) and decision trees on predicting ear conditions with feature extracted by filter bank, discrete cosine transform (DCT) and color coherence vector (CCV)~\cite{external}. More recently, deep learning has been applied to classify otoscopy images. Zafer studies the combination of the fused fine-tuned deep features and SVM model applied to 3 diagnostic class image classification~\cite{finetune}.

Furthermore, Khan et al.~\cite{tympanic} compare the performance of a variety of CNN architectures and find that DenseNet~\cite{huang2017densely} produces the best result in the classification of 3 diagnostic classes. The authors also use Grad-CAM~\cite{selvaraju2017grad} to visualize the important regions in the image for the model. 

In~\cite{ensemble}, the authors compare the performance applying 9 kinds of ImageNet~\cite{deng2009imagenet} pretrained CNN networks to 6 class otoscopy image classification and then further improve the performance by ensembling the output of 2 networks.
The prementioned deep learning based methods overall show great performance in the respective settings, achieving high accuracy ranging from 94\% to 99\%.

The accuracy reported by previous work, even in multi-class settings, is remarkable, yet, all previous approaches process high quality still images of the eardrum. Unfortunately, in a clinical setting the capture of high quality still images can be challenging in practice, especially given that pediatric patients are often moving and uncooperative with examination. The use of a single image also increased the chance of failing to capture a clinically accurate representation of the anatomy due incomplete view.  A method that analyzes otoscopy video sequences rather than still images could help overcome aforementioned shortcomings. 

The straightforward approach to have the whole video sequence as input is to train a 3D (2D images plus temporal dimension) CNN mapping from videos to labels. However, that would require 1) heuristics to harmonize videos to a preset length, and 2) large amounts of labeled data to avoid over-fitting, especially considering that the class frequency in any otoscopy video dataset is likely to be heavily skewed towards normal cases. The present work addresses these challenges by framing otoscopy interpretation as a video anomaly detection problem. Under our setting, the model is trained with normal videos only and will flag videos that diverge from the normal as an anomaly during testing. 

The contribution of this work is two-fold. First, we devise a two stage model to apply on full video sequences collected in clinics, which differentiates our approach from all previous methods that only considered high quality still images. A region detector first extracts eardrum patches from video frames that are then used to perform anomaly detection in the following stage. This architecture restricts anomaly detection to semantically useful regions. Second, we develop the shift contrastive anomaly detection (SCAD) method that leverages color-jitter based distributional shift-transformation to improve the separability between normal and abnormal data. Tailored to our application, our self-supervision task enforces the model to leverage subtle color features to identify abnormality during testing. Even when the amount of training videos is relatively small, our approach generates superior binary (normal / abnormal) otoscopy video screening results compared to both a baseline method and the average of a group of clinicians. We believe that this work constitutes a promising first step towards achieving algorithmic decision support for the diagnosis of pathology of the middle and external ear, and the development of smart tool which might aid clinicians in the accurate diagnosis and management of common ear disease.

\section{Related work}

\IEEEPARstart{T}{he} task of anomaly detection~\cite{chalapathy2019deep, ruff2021unifying}, also referred to as out-of-distribution or novelty detection depending on context, is to identify unfitted data samples. This is a crucial task in many real world applications such as detecting system malfunction, financial fraud and health issues. Anomaly detection is a useful but challenging task, because anomalies are hard to define. In a supervised anomaly detection setting, abnormal samples are provided to the model in the training stage to give guidance on what qualifies as an "anomaly". However, there could be many reasons a data sample is abnormal such that collecting a representative amount of abnormal samples is a hard problem itself. This is especially challenging in medical applications, where the dataset is very likely to heavily skewed towards normal class. Unsupervised anomaly detection (UAD), on the other hand, does not need normal / abnormal supervision signal during training. Current approaches usually utilize only the normal data during training, which makes it more appealing to applications with skewed dataset. Therefore, in the scope of this paper, we focus only on the UAD setting where the detector can only access to samples from the normal data distribution during training. Moreover, in many real world applications, the task is perform UAD on high-dimensional imagery data (e.g., in our case, detecting anomaly from endoscopic videos), which makes the problem even more challenging.

A rich set of methods~\cite{chalapathy2019deep} has been proposed to study UAD problem on imagery data, approaching the problem with four main paradigms: i) Density-based methods first estimate the normal data density using methods including Gaussian Mixture and Energy Based Models, and then detect data with low estimated density as anomaly. DAGMM~\cite{zong2018deep} and DSEBM~\cite{zhai2016deep} are methods that belongs to this category. ii) One-class classifier-based methods fit a classifier, e.g., Deep SVDD~\cite{pmlr-v80-ruff18a} and DROC~\cite{sohn2020learning}, to separate normal and all other data and then use it to detect anomalies. iii) Reconstruction-based techniques learn a reconstruction model, e.g., AnoGAN~\cite{schlegl2017unsupervised}, of normal images and detect anomalies as samples with high reconstruction error. iv) Self-supervised-based methods learn a feature extractor using self-supervised tasks, such as distinguishing whether or not certain transformation has been applied to the image~\cite{kolesnikov2019revisiting}. Then, during learning features of augmented versions of the same image should be closer than features of different images~\cite{chen2020simple}. Upon convergence, anomaly detection is performed on the extracted features. Notable methods along this line include SVD-RND~\cite{choi2019novelty}, CutPaste~\cite{li2021cutpaste}, CSI~\cite{CSI}, SSD~\cite{sehwag2021ssd} and MSC~\cite{MSC}. UAD has also been applied to medical imaging~\cite{fernando2020deep} across many domains, including X-ray~\cite{davletshina2020unsupervised, bozorgtabar2020anomaly}, CT~\cite{sato2018primitive, pawlowski2018unsupervised}, MRI~\cite{baur2021autoencoders, han2021madgan, baur2020steganomaly} and endoscopy~\cite{liu2019unsupervised} datasets. 

Anomalies in medical imaging tend to be more subtle and only reside in a small region of the whole image, which makes it particularly challenging to apply UAD in this field. To address this problem, ~\cite{zimmerer2018context} proposes to connect Context-encoding (CE) and Variational Autoencoders (VAE) to combine both reconstruction and density based anomaly score. Doing so shows superior performances in three brain MRI image datasets. To directly target subtle anomaly in small regions, there are efforts to develop self-supervised tasks that are tailored to a certain anomaly group. It is desirable to apply artificial transformation of data that is similar to real anomaly so that the self-supervised task can force the model to learn features that are discriminative enough to detect anomaly in testing. As an example, CutPaste~\cite{li2021cutpaste} creates discontinuous defects by cutting a region of an image and pasting it to another location. While this method works well for industrial visual inspection, the introduced sharp discontinuity is uncommon in medical imagery making this method less useful for anomaly detection in medical data. Alternatively, foreign patch interpolation (FPI)~\cite{tan2020detecting} creates artificial defects by interpolating a small local patch with a foreign patch of the image. A network is trained to estimate the pixel-wise interpolating factor from the synthesised image. The estimated interpolating factor is then used as anomaly score during testing. Success is demonstrated using Brain MRI and abdominal CT image datasets. Poisson image interpolation (PII)~\cite{tan2021detecting} extends this idea by using Poison image editing to blend in the foreign patch instead of direct linear interpolation, targeting more subtle and continuous irregularities. Improvement in performance is shown using chest X-ray and fetal ultrasound datasets. Meanwhile, both FPI and PII utilize self-supervised tasks targeted at regional shape anomaly which is less applicable to our problem where subtle color anomaly is more crucial.

\section{Methodology}
\begin{figure*}
  \begin{center}
  \includegraphics[width=6.5in]{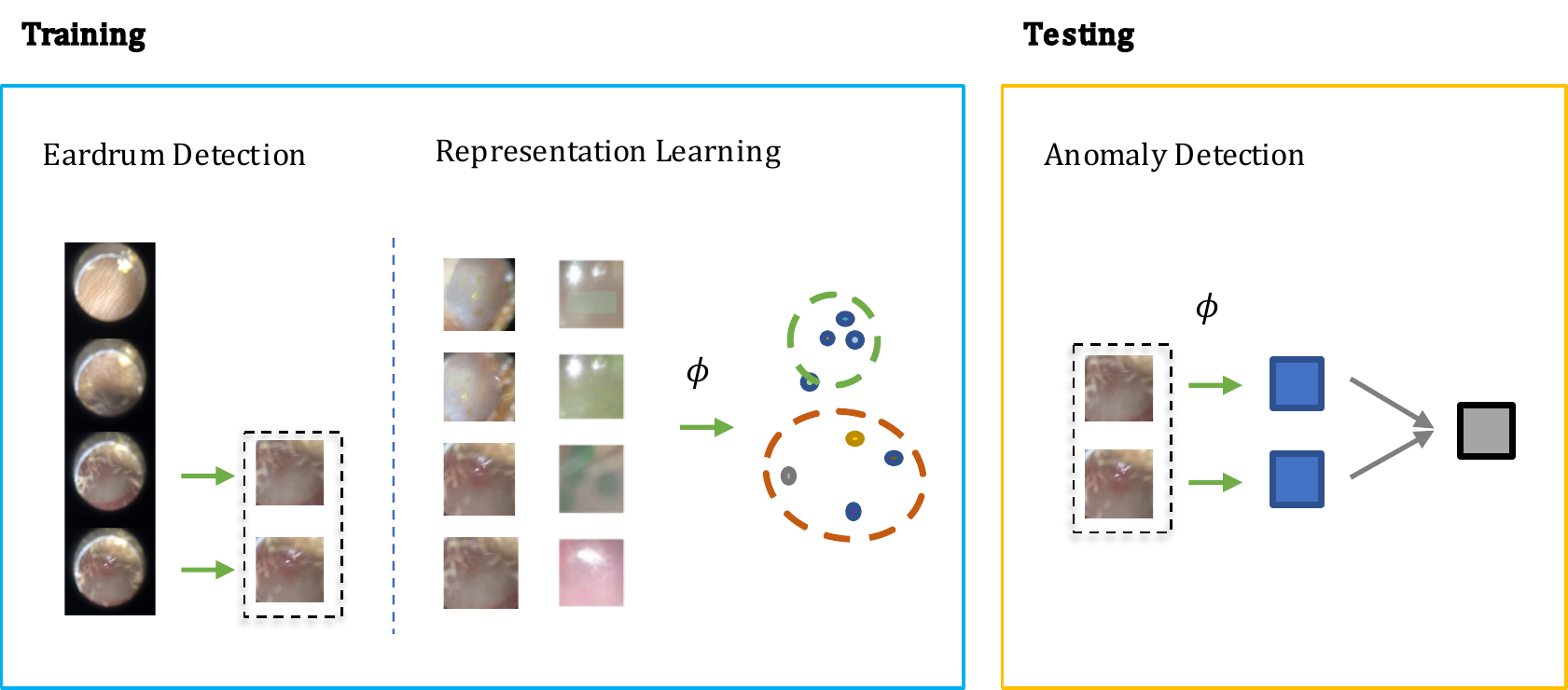}\\
  \caption{The screening architecture of our shift contrastive anomaly detection (SCAD) is composed of three sub-blocks: eardrum detection, representation learning and anomaly detection. During training, eardrum detection is learned by supervised training with provided labels. Representation learning is conducted with self-supervision to obtain an embedding function $\phi$. During testing, we use the embedding function $\phi$ to produce frame level anomaly score and video level anomaly score to perform anomaly detection.}\label{fig:architecture}
  \end{center}
\end{figure*}

\IEEEPARstart{T}{he} otoscope video database is skewed towards the non-disease status given the practice of capturing bilateral ear examinations and occurrence of non-ear disease presenting with symptoms such as ear pain. Dataset imbalance such as this can present significant challenge for the development of robust design boundaries, particularly when relying on small data-sets. To overcome such limitations, we propose use unsupervised anomaly detection approach to provide computer-aided screening for otoscopy video. The formal definition of this problem is as follows:

We are given a large training dataset $\mathcal{D}_{\emph{train}}$ comprising only normal video sequences and a smaller testing dataset $\mathcal{D}_{\emph{test}}$ comprising both normal and abnormal video sequences. Given the training dataset, our objective is to learn an anomaly score function $A$(${v}$) with video $v$ as input, and detect the abnormal videos as anomaly during testing. Ideally, the function learns to map normal videos to small anomaly scores and abnormal videos to large anomaly scores. During testing, we threshold the score, where $A$($s$) $ > \psi$ indicates anomaly.

In otoscopy video diagnosis, the temporal relation between frames is not as influential as in action recognition tasks. Whether or not a video is abnormal is directly determined by the existence of abnormal frames therein. With this observation, we turn the video anomaly detection problem into a frame anomaly detection problem with an additional video level aggregation function. Moreover, only some of the frames in the video are vital to diagnosis, and in those frames, only the region that visualizes the eardrum. Therefore, it would be desirable to only conduct anomaly detection on such frames, and the eardrum region in particular. As shown in Figure~\ref{fig:architecture}, our method has two steps during training: Step 1) Supervised learning of a detection module that extracts eardrum patches $x$ from given video $v$ using provided labels. Step 2) Self-supervised representation learning on the labeled region of interest to obtain an embedding function that ideally projects normal and abnormal samples to distinct regions of the embedding space. During testing, we compute anomaly scores using the detected patches and the learned embedding function. 

\subsection {Eardrum detection}
To detect the frames and regions showing the eardrum, we use a convolutional neural network and train it in supervised manner using image-label pairs. Different from the standard detection problem where each image may have multiple instances, we can only have at most one eardrum in each frame. Therefore, we do not use a standard object detection architecture, and rather choose to have a single convolutional neural network to predict the binary label and corresponding bounding box. We define $\mathcal{L}_{\emph{cls}}$ as the cross entropy loss of groundtruth label and the classification score, and define $\mathcal{L}_{\emph{local}}$ as the $L1$ loss of the groundtruth bounding box and the predicted bounding box. Then, the overall loss is simply combining the classification loss and localisation loss.

\begin{equation}
    \mathcal{L}_{\emph{detect}} = \mathcal{L}_{\emph{cls}} + \mathcal{L}_{\emph{local}}
\end{equation}

Given our problem setup, we train the detection module on the normal training videos only. Even though the model does not use abnormal videos during training, we find it generalizes well to abnormal frames during testing as shown in section~\ref{detection_result_section}. 

\subsection {Shift Contrastive Anomaly Detection (SCAD)}
\begin{figure*}
  \centering
    \includegraphics[width=6.75in]{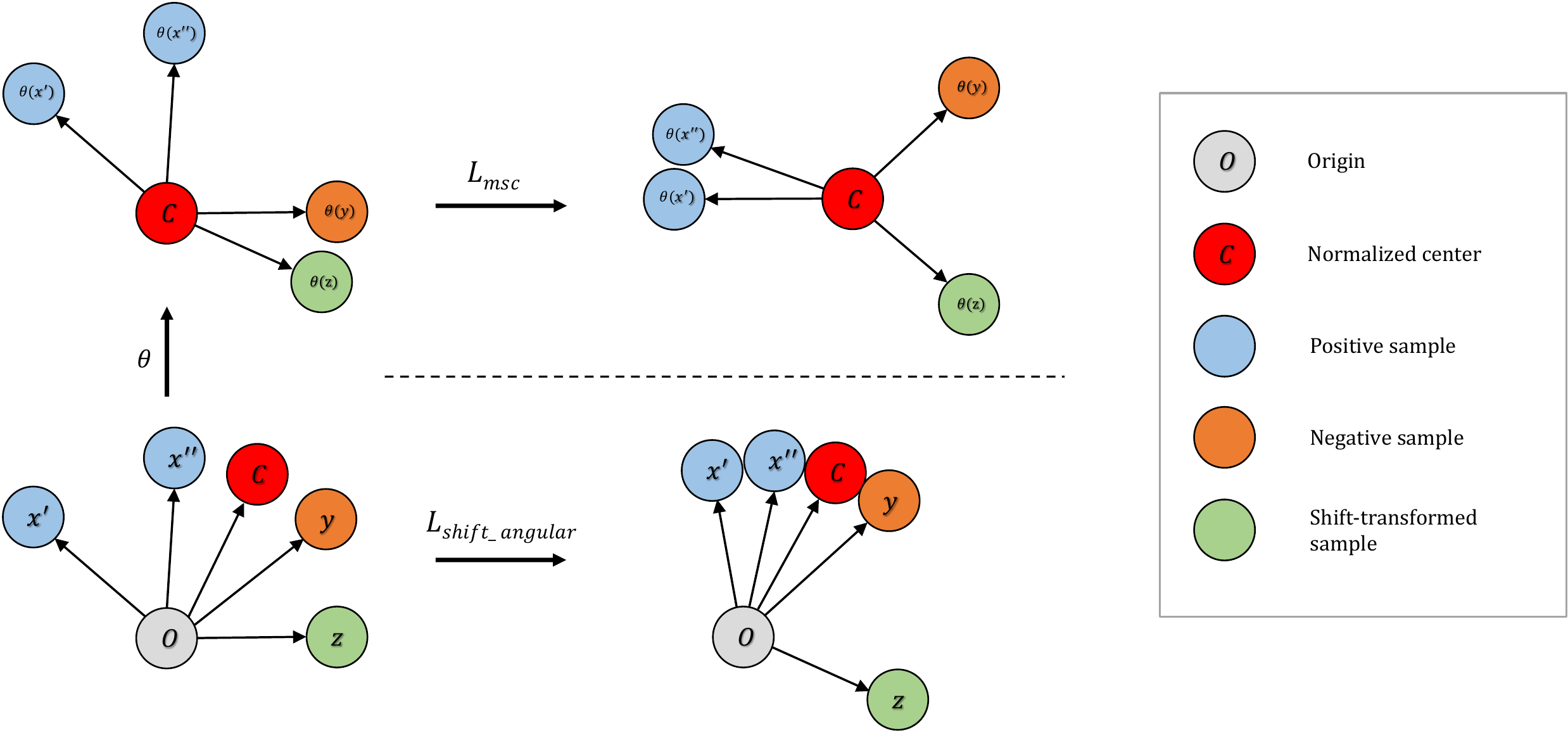}
    \caption{\textbf{Top:} On the mean-shifted representation space, $\mathcal{L}{\emph{msc}}$ maximizes the angle between negative pairs. Shift-transformed samples are treated as additional instances in this step. \textbf{Bottom:} On the angular representation space $\mathcal{L}{\emph{shift\_angular}}$ increases separability of anomaly by i) increasing the angle between the embedding of normalized center and the shift-transformed samples; ii) decreasing the angle between embedding of normalized center and normalized samples.}
  \label{fig:methods}
\end{figure*}

\textbf{Contrastive Learning:} Contrastive learning~\cite{chen2020simple,MOCO} has become the top performing self-supervised learning method in recent years. In this paradigm, the first step of the training procedure is to sample a minibatch of size $N$, and perform augmentation twice to each sample $x_i$ to obtain $(x_{i^{\prime}}, x_{i^{\prime\prime}})$ termed as positive pair, producing $2N$ samples in total. All images are passed through a feature extractor and then features are typically scaled to the unit sphere by $l_2$ normalization to get representation $\phi$. The contrastive loss to pair $(x_{i^{\prime}}, x_{i^{\prime\prime}})$ is then defined as
\begin{equation} 
    \mathcal{L}_\emph{con}(x_{i^{\prime}}, x_{i^{\prime\prime}}) = -\log{
        \frac{\exp(\phi({x_{i^\prime}})\cdot{\phi(x_{i^{\prime\prime}})}/\tau)}
            {\sum_{m=1}^{2N}{\mathbbm{1}[i^{\prime} \neq m] \cdot
                \exp(\phi({x_{i^\prime}})\cdot{\phi(x_m)}/\tau)
                }
            }
    }\,,
    \label{eq:l_con}
\end{equation}
where $\tau$ is a temperature hyper-parameter.

The contrastive learning objective pulls $x_{i^\prime}$ close to $x_{i^{\prime\prime}}$ and pushes all other samples away from $x_{i^\prime}$. Intuitively, this will make the learned embedding capture a latent structure that is meaningful enough to separate samples from one another and thus be beneficial for downstream tasks. This paradigm shows great success in self-supervised training for image recognition, but has inherent problems for anomaly detection. To minimize the loss function, the angles between positive and negative samples need to be maximized even though both samples are from the normal class. This results in a scenario where representations of normal samples span across the whole unit sphere. During testing, anomaly sample could therefore be projected onto a location that is close to normal embeddings, which challenges the paradigm of anomaly detection.
\begin{figure*}
  \centering
    \includegraphics[width=6.75in]{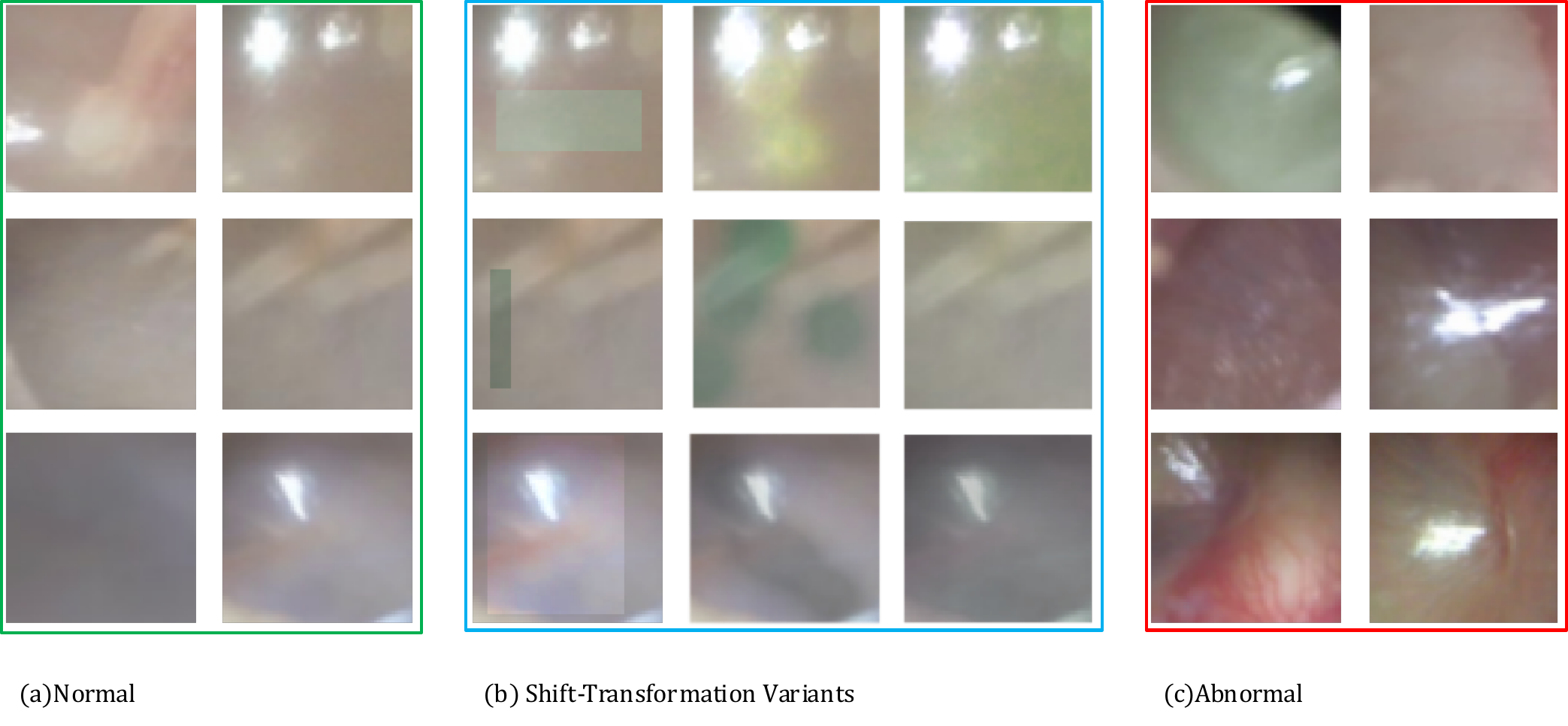}
    \caption{\textbf{(a)} Normal eardrum region examples. \textbf{(b)} Shift-transformation variants applied to the second normal example of each row. From left to right: color-jitter-random-cut (CJ-RC), color-jitter-random-region (CJ-RR), color-jitter-whole-frame (CJ-WF). \textbf{(c)} Abnormal eardrum region examples.}
  \label{fig:augmentations}
\end{figure*}

\textbf{Mean-Shifted Contrastive Loss:} To adapt contrastive learning to anomaly detection, MSC~\cite{MSC} proposes the alternative mean-shifted contrastive loss. Rather than directly minimizing the contrastive loss in the representation space, MSC constructs a mean-shifted counterpart, by subtracting the center $c$ of the whole training set and then normalizing to the unit sphere. For a given sample $x$, the mean-shifted embedding is defined as 
\begin{equation}
    \theta(x) = \frac{\phi(x) - c}{\norm{\phi(x) - c}}\,.
    \label{eq:l_angular}
\end{equation}
The mean-shifted loss is then constructed by applying typical contrastive loss on this mean-shifted embedding space: 

\begin{equation}
    \mathcal{L}_\emph{msc}(x^{\prime},x^{\prime\prime}) =  \mathcal{L}_{\emph{con}}(\theta(x^{\prime}), \theta(x^{\prime\prime}))\,.
    \label{eq:l_shift_angular}
\end{equation}

Because the mean-shifted representation is normalized around $c$, the training no longer spans the normal samples across the whole unit sphere of $\phi(x)$. This makes anomaly samples more separable from normal samples in the $\phi(x)$ space.

\textbf{Distributional shift angular loss:} To further improve separability, MSC~\cite{MSC} uses additional angular center loss
\begin{equation} \label{msc_loss}
    \mathcal{L}_{\emph{angular}} = -\phi(x)\cdot c\,,
\end{equation}
to shrink normal samples around the normalized center. The assumption is that normal data lying in a small region around the center will be more discriminative. The total loss that MSC uses then becomes $\mathcal{L}_{\emph{angular}} + \mathcal{L}_{\emph{angular}}$.

This approach works well in datasets composed of regular images (e.g., CIFAR dataset~\cite{krizhevsky2009learning}), where images from different classes are treated as anomaly. Empirically we find that this strategy does not perform as well in our medical imaging application. The main reason is that we are having much smaller semantic variations between normal and abnormal images so that their respective embeddings are still close to each other even after training using mean-shifted contrastive and angular center loss. 

To address this issue, we develop the distributional shift angular center loss to further enhance separability between normal and abnormal samples. Our method constructs additional shift-transformed samples $z$ from $x$ and then pushes $\phi(z)$ away from the normalized training center $c$. Shift-transformed samples $z$ are created by applying distributional shift-transformations on $x$ so that $z$ are no longer from the original data distribution. The intuition is straightforward: not only should the normal samples occupy a small region around the center, but also the samples not drawn from the original distribution cannot be too close to the center either. Hinge loss is applied to the angle between center $c$ and shift-augmented samples $z$. The loss function takes the form:

\begin{equation}
    \mathcal{L}_{\emph{shift\_angular}} = -\phi(x)\cdot c + max(0, 1-\phi(z)\cdot c)\,.
\end{equation}

Ideally, these shift-transformed samples $z$ are semantically similar to real anomalies. In the context of otoscopy images, color is an important feature in that infection may lead to change of color of the whole or part of the eardrum. Thus, we employ three variants of color-jitter as the distributional shift-transformation. They are color-jitter-random-cut (CJ-RC), color-jitter-random-region (CJ-RR), and color-jitter-whole-frame (CJ-WF). CJ-RC uses a random rectangle as a mask of transformation. CJ-RR is constructed by interpolating the color-jittered image and the original image using pixel-wise weights for each image. We first initialize the pixel-wise weights to be all zero, and then pick random points to be filled with one, and apply Gaussian filter to obtain smoothed weights. CJ-WF simply applies color-jitter to the whole frame. Figure~\ref{fig:augmentations} presents illustrations of the three augmentation strategies.

\textbf{Final loss:} We combine i) the mean-shifted contrastive loss $\mathcal{L}_{msc}$ and ii) the angular loss with distributional shift augmentations $\mathcal{L}_{shift\_angular}$: 
\begin{equation}
	\mathcal{L}_{final} = \mathcal{L}_{msc} + \mathcal{L}_{shift\_angular}\,.
	\label{eq:final_loss}
\end{equation}
This loss function is tailored to this specific medical imaging application and enjoys the best of both objectives. The mean-shifted loss makes the features to be representative of the images and the angular loss encourages normal and abnormal instances to be distant to each other in the feature space. Therefore, the combination of these two losses achieves superior performances, which is demonstrated through ablation in section~\ref{scad_result}. 

\textbf{Frame level anomaly score:} In order to classify a sample as normal or abnormal, we use a simple criterion based on k-Nearest-Neighbors (kNN). kNN predicts the label of a data point by ensembling the k closest labeled samples according to some distance measure. In this work, we use the cosine distance between the features of the target image $x$ and those of all training images. The anomaly score is then given by:

\begin{equation} \label{frame_score}
a(x) = \Sigma_{\phi(y) \in N_k(x)}(1 - \phi(x) \cdot \phi(y))\,,
\end{equation}

where $N_k(x)$ denotes the $k$ nearest features to $\phi(x)$ in the training feature set $\{\phi(z)\}_{z \in \mathcal{X}_{train}}$.

\textbf{Video level anomaly score:} We consider a relatively simple method to aggregate the frame level anomaly score to construct the video level anomaly score. The aggregation function is simply taking the average of the frame level anomaly score across all detected eardrum patches. The video level anomaly score is defined as 
\begin{equation} \label{video_score}
A(s) = \overline{\{a(x)\}}\,,
\end{equation}
where $\{a( x)\}$ represents the set of frame level anomaly scores for all frames in the given video $ s$.

In summary, during training, we learn a frame level embedding function $\phi$ with all individual normal frames in the training set. During testing, we compute the kNN-based anomaly score using the embedding feature from $\phi$ for all detected frames. We then aggregate the predicted anomaly score for each detected frame in testing video to produce a video level anomaly score $A$ through averaging. Such a score is expected to be small for normal videos and large for abnormal videos. Any frame that is substantially different from normal frames in the training set (e.g., a red, bulging eardrum) would lead to a large distance to training examples in embedding space and then produce large image and video level anomaly score. The threshold can be selected to meet desired clinical requirement (e.g., a certain specificity value) and in practice is selected using the validation dataset consisting of both normal and abnormal videos. 

\begin{figure*}
  \centering
    \subcaptionbox{}{
        \includegraphics[width=7in]{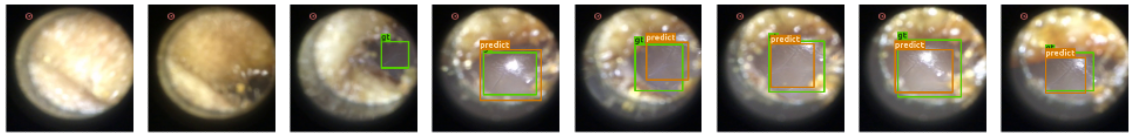}
    }
    \subcaptionbox{}{
        \includegraphics[width=7in]{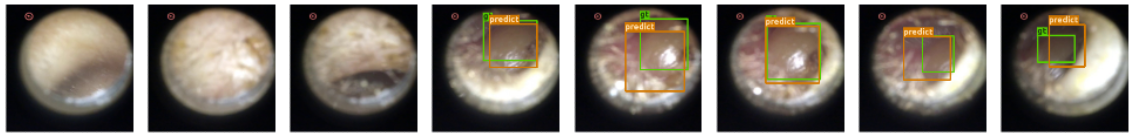}
    }
    \caption{Our detection module successfully extracts eardrum regions in video sequences. Green bounding box represents groundtruth eardrum region; orange bounding box represents detected eardrum region. \textbf{Top Row:} Normal video example. \textbf{Bottom Row:} Abnormal video example.}
  \label{fig:detection_vis}
\end{figure*}

\section{Experimental results}

\subsection{Data preparation}

We collected a total of 100 otoscopy videos from pediatric patients that were seen for various conditions in an urban, tertiary-care pediatric emergency department that treats over 34,000 patients, aged between 0 and 21 years of age per year. Patients were recruited as a convenience sample based on willingness to participate in the study. At this site patient ages range from 0-22 years. Study protocols were approved by the local institutional review board. The length for each video ranges from 5 to 40 seconds. Twenty seven to thirty frames per second are extracted from the video. Two otolaryngologists (ears, nose, throat specialists) annotated every video and consensus agreement was used as ground truth. While the expert annotations are more granular, due to the small size of the dataset, we decided to restrict analysis to two classes and each video is assigned into either the normal or abnormal category. Additional to binary categorical label on video level, we also annotate frame level binary labels and bounding boxes for each appearance of the eardrum, where the eardrum occupies at least 20\% of the image size in width and height. Of the 100 videos, 80 are labeled normal and 20 are labeled abnormal. We assign 60 normal videos to the training set, 10 to the validation set and 10 to the testing set. For anomaly videos, we assign 10 to the validation set and 10 to the testing set. The testing set is used for evaluation of the algorithms and comparison with human clinicians. We do not have overly apparent abnormal examples (e.g., the whole eardrum is filled with blood) during testing which makes it a challenging comparative study with human clinicians. 

Video capture of ear exam was performed using Cellscope Oto attached to a first-generation iPhone SE. Maintaining relative uniformity of video settings such as color are critical to the accuracy of the AI algorithm. Clinical deployment of such technology therefore presents significant challenge given the ad hoc selection of otoscopes used in the clinical setting and perhaps more importantly the overwhelming reliance on non-digital otoscopes in clinical practice. As a result we opted to use Cellscope Oto as an affordable solution that can potentially be deployed at scale. The Cellscope Oto device is designed to attache to an ear speculum and provides an additional light source but relies on the iPhone camera to capture the exam. We do note that the image and video quality captured on this device is not as high as reported in previous works~\cite{external, prototype}, especially with respect to as sharpness and clarity. Unlike Cellscope oto however more advanced high quality otoscopes with digital video capacity (as used in \cite{external, prototype}) are often cost prohibitive to wider adoption. 

\subsection{Eardrum detection} \label{detection_result_section}
We first train a model for eardrum detection to extract eardrum patches. We use Resnet-101 as backbone network and choose the stochastic gradient descent (SGD) optimizer with $1e-3$ learning rate, $0.9$ momentum and $1e-4$ weight decay. The model is trained using all frames from the normal training set videos for 5000 epochs with a batch size of 128. The input images are resized to 256 by 256 and randomly cropped to 224 by 224, followed by a random selection of data augmentation of color-jitter, cutout, rotation and shearing. 

\textbf{Results:} Since under our setup there is at most one instance of a single class, we evaluate the performance of our model by accuracy of eardrum detection under different intersection over union (IoU) \cite{rezatofighi2019generalized} thresholds. A correct detection is defined as a frame that produces a correct frame-level classification and, if the given frame shows the eardrum, also produces a bounding box with IoU score larger than a certain threshold in that given frame. As shown in Table~\ref{table:detection}, our model produces high accuracy across varied IoU thresholds. A visualization of eardrum detection in video sequences can be found in Figure~\ref{fig:detection_vis}. Note that even though the model is supervised only on instances from training normal videos, it generalizes well to both normal and abnormal videos in the testing set. This indicates that the network extracts features that are more consistent to the image semantic and agnostics to the details on eardrum. 

\begin{table}[ht]
    \centering
    \begin{tabular}[t]{l c c c c} 
        \toprule
        {} & {Accu (IoU$>$.50)}&{Accu (IoU$>$.75)}&{Accu (IoU$>$.90)}   \\ 
        \midrule
        {Normal} & $77.3$ & $69.7$ & $37.8$\\
        {Abnormal} & $82.1$ & $79.3$ & $34.4$\\
        {Overall} & $80.8$ & $73.1$ & $35.7$\\
        \bottomrule
    \end{tabular}
    
    \caption{Detection performance evaluated by accuracy~\% under different IoU thresholds. }
    \label{table:detection}
\end{table}

\subsection{Anomaly detection} \label{scad_result}

Following the architecture suggested in~\cite{MSC}, we construct the feature extractor $\phi$ to be the ResNet-101 convolutional neural network followed by an additional l2 normalization layer. We augment the normal samples by sequentially applying 224 by 224-pixel crop from a randomly resize image and random horizontal flips. The model is initialized with the ImageNet pre-trained model and then the two last blocks of the ResNet-101 are fine-tuned for 5000 epochs with the loss function in Equation~\ref{eq:final_loss}. The temperature $\tau$ in \ref{eq:l_con} is set to be 0.25. We use SGD optimizer with learning right of $1e-5$ weight decay of $5e-5$, and no momentum. For each minibatch size of 60, we sample one frame from each video to prevent similar frames treated as negative pairs. During training, we use the groundtruth bounding box to extract patches from videos. We use KNN (k=2) in computing the frame-level anomaly score. We conduct experiments with three random seeds for evaluations.

\begin{figure*}
  \centering
  \subcaptionbox{MSC.}{
    \includegraphics[width=1.4in]{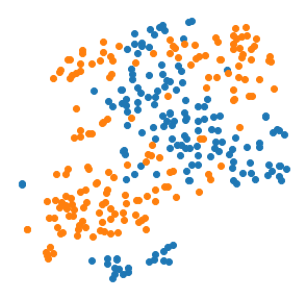}
  }
  \subcaptionbox{AMSC-CJ-RC.}{
    \includegraphics[width=1.4in]{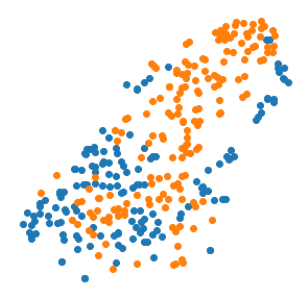}
  }
    \subcaptionbox{AMSC-CJ-RR.}{
    \includegraphics[width=1.4in]{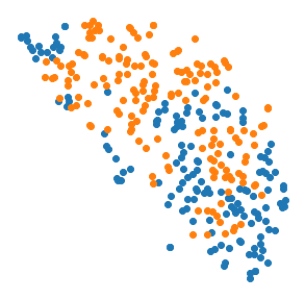}
  }
    \subcaptionbox{AMSC-CJ-WF.}{
    \includegraphics[width=1.4in]{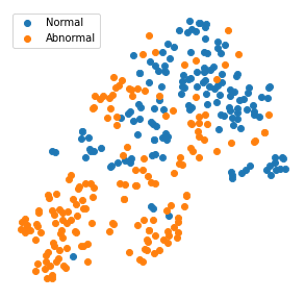}
  }
    \caption{Visualization of the embedding for all frames in the testing set of normal and abnormal videos using t-SNE.}
  \label{fig:embedding}
\end{figure*}

\begin{figure*}
  \centering
  \subcaptionbox{MSC.}{
    \includegraphics[width=1.4in]{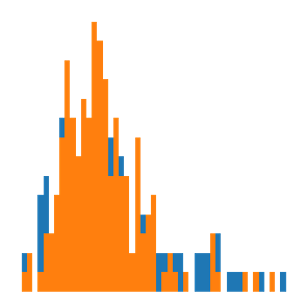}
  }
  \subcaptionbox{AMSC-CJ-RC.}{
    \includegraphics[width=1.4in]{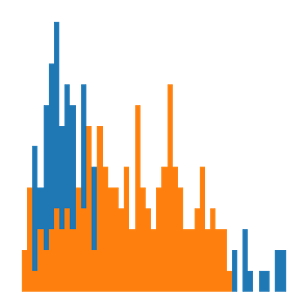}
  }
    \subcaptionbox{AMSC-CJ-RR.}{
    \includegraphics[width=1.4in]{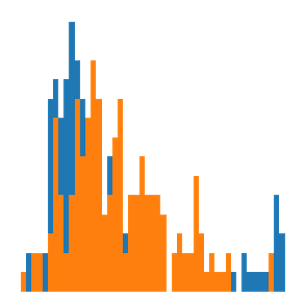}
  }
    \subcaptionbox{AMSC-CJ-WF.}{
    \includegraphics[width=1.4in]{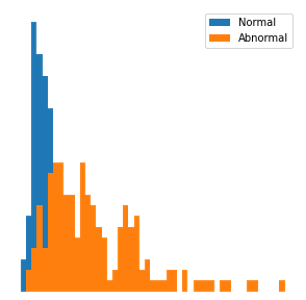}
  }
    \caption{Distribution of the frame-level anomaly score for all frames in the testing set of normal and abnormal videos.}
  \label{fig:distribution}
\end{figure*}

\textbf{Results:} We evaluate our methods on the anomaly detection task in the previously introduced otoscopy video dataset. We adopt the Area Under the Receiver Operating Characteristic curve (AUROC) and the Area Under the Precision Recall Curve (AUPRC) as anomaly detection performance score. We compare our approaches against the recently proposed top performing method MSC~\cite{MSC}. The training and testing procedures are the same as ours except that it does not utilize shift-transformations and uses the angular loss~(\ref{eq:l_angular}) instead of shift angular loss~(\ref{eq:l_shift_angular}) in computing the final loss function. In Table~\ref{table:auroc_auprc}, we present a comparison between MSC and three variants of SCAD evaluated by AUROC and AUPRC. We see that the SCAD-CJ-WF variation performs the best, reaching an AUROC of 88.0\% and AUPRC of 87.9\% on average. This suggests that i) color is indeed a very important feature in separating anomaly from normal in otoscopy images, and that ii) color jitter on the whole image can produces semantically more similar samples to the real anomaly. This could be due to that both random cut and random region introduce artificial patterns on the image that are utilized in the self-supervised training but are not seen in the abnormal samples during testing. To intuitively visualize the feature embedding produced by different methods, we plot the latent embedding for all testing frames in Figure~\ref{fig:embedding} and the corresponding kNN based frame-level anomaly score in Figure~\ref{fig:distribution}. As we can see, SCAD-CJ-WF generates the embedding that is most separable both qualitatively and quantitatively as measured by the anomaly score.
 
\begin{table}[ht]
    \centering
    \begin{tabular}[t]{l c c c c} 
        \toprule
        {Method} & {AUROC} &  {AUPRC}   \\ 
        \midrule
        {MSC} & $64.3 \pm 4.0$ & $60.7 \pm 2.4$ \\
        {SCAD-CJ-RC} & $78.0 \pm 1.0$ & $69.1 \pm 1.1$ \\
        {SCAD-CJ-RR} & $82.3 \pm 1.2$ & $74.4 \pm 1.1$ \\
        {SCAD-CJ-WF} & $\mathbf{88.0 \pm 1.0}$ & $\mathbf{87.9 \pm 1.8}$ \\
        \bottomrule
        
    \end{tabular}
    
    \caption{Results evaluated by AUROC \% and AUPRC \%. }
    \label{table:auroc_auprc}
\end{table}

\textbf{Training objective:} The individual effect of each loss component is presented in Table~\ref{table:ablation}. We note that neither $\mathcal{L}_{\emph{msc}}$ or $\mathcal{L}_{\emph{angular}}$ individually performs well in our dataset. $\mathcal{L}_{\emph{shift\_angular}}$ outperforms all other individual objective and combining it with $\mathcal{L}_{\emph{msc}}$ results in further improvement.

\begin{table}[ht]
    \centering
    \begin{tabular}[t]{l c c c c} 
        \toprule
        {Metric} & {$\mathcal{L}_{\emph{msc}}$} &  $\mathcal{L}_{\emph{angular}}$ & $\mathcal{L}_{\emph{shift\_angular}}$ & $\mathcal{L}_{\emph{msc}}$ + $\mathcal{L}_{\emph{shift\_angular}}$ \\ 
        \midrule
        {AUROC} & $59.3 \pm 0.6$ & $57.7 \pm 4.0$ & $78.0 \pm 1.0$ & $\mathbf{88.0 \pm 1.0}$\\
        {AUPRC} & $55.1 \pm 1.6$ & $54.5 \pm 3.0$ & $69.8 \pm 0.8$ & $\mathbf{87.9 \pm 1.8}$\\

        \bottomrule
        
    \end{tabular}
    
    \caption{Training objective ablation study (AUROC \% and AUPRC \%).  $\mathcal{L}_{\emph{shift\_angular}}$ is of the SCAD-CJ-WF variant. Note that the MSC method uses $\mathcal{L}_{\emph{msc}}$ + $\mathcal{L}_{\emph{angular}}$ as loss function and the result is presented in Table~\ref{table:auroc_auprc}}
    \label{table:ablation}
\end{table}

\subsection{Comparison with human clinicians}
To further evaluate our performance in a real-world setting, we performed a cross-sectional study comparing the performance of clinicians versus our model. The study was approved by the local institutional review board.

\textbf{Study Setting:} 
Clinicians who routinely evaluate eardrums by otoscopy in a primary care, urgent care, or emergency medicine setting were invited to participate in this study. Qualifying clinicians included physicians (MD/DO), nurse practitioners (NP), and physician assistants (PAs). These clinician group reflects frontline healthcare providers who typically diagnose and manage ear disease. Exclusion criteria were practicing outside of the United states,  employment by the Johns Hopkins University or Johns Hopkins Hospital, visual impairment limiting video analysis, or non-English speakers. Recruitment took place through digital messaging via professional email mailing lists and professional social media groups during November 2020.

\textbf{Study Procedure:} 
Ten normal and ten videos designated as acute otitis media (AOM) by ground truth were selected as the testing set of images, as described above. Clinicians completed an online servey in which they were instructed to review the 20 eardrum videos without any further clinical information provided. Following each video, participants were asked to indicate whether the eardrums appeared normal or abnormal. They were also asked to rate their level of confidence in the accuracy of their diagnosis using a Likert scale (1= not at all confident to 4 = very confident). Participants were also asked to report the number of times they viewed each video. Participant responses were collected and managed using REDCap electronic data capture tools~\cite{redcap} hosted at Johns Hopkins University.

\begin{figure}
  \centering
    \includegraphics[width=3.2in]{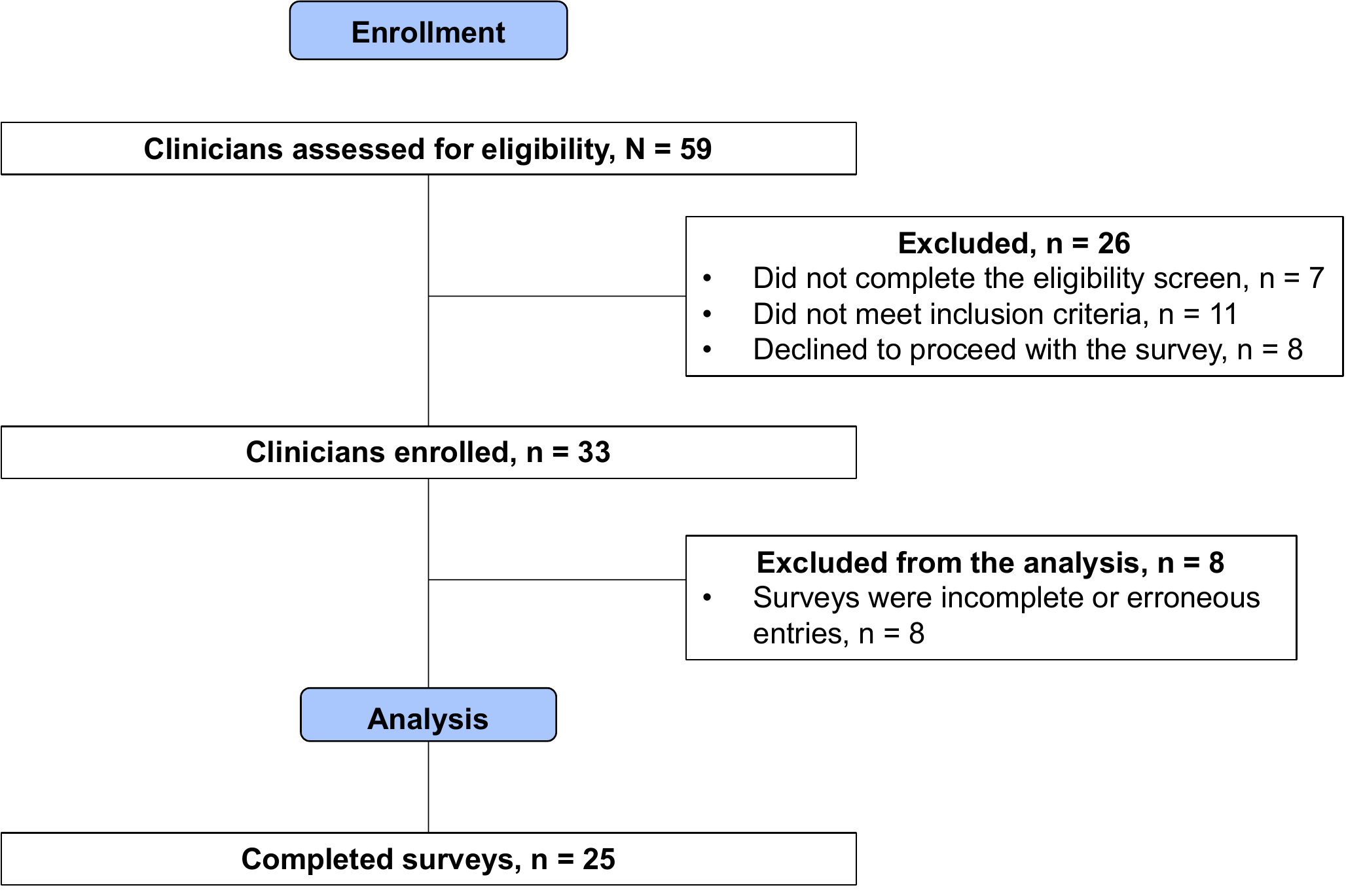}
    \caption{Consort flow diagram of clinician enrollment. After removing ineligible samples, 25 surveys were included for comparative analysis.}
  \label{fig:consortflow}
\end{figure}

\begin{table}[ht]
\centering
\begin{tabular}[t]{l c c c} 
    \toprule
    {Characteristic} & {Sample Size} & {Accuracy}\\ 
    \midrule
    {\textbf{Sex}} &  &\\
    {   Female} & 15 (60\%)& $73.0 \pm 15.1$ \\
    {   Male} & 10 (40\%)& $60.0 \pm 17.5$ \\
    \midrule
    {\textbf{Current Practice}} &  &\\
    {   General Pediatrics} & 12 (48\%)& $73.0 \pm 15.1$\\
    {   Pediatric Specialty} & 5 (20\%)& $70.0 \pm 22.4$\\
    {   General Internal Medicine} & 3 (12\%)& $55.0 \pm 8.7$\\
    {   Others} & 5 (20\%) & $67.0 \pm 8.4$\\
    \bottomrule
\end{tabular}
\caption{\label{table:demographics} Clinician demographics and corresponding accuracy (mean \% and standard deviation \%) by groups. Other practices include emergency medicine, family medicine and urgent care.}

\end{table}

\textbf{Study Population:}
A total of 59 clinicians were recruited and assessed for eligibility. After removing ineligible, declined, or incomplete surveys, 25 surveys were included for analysis. The process is shown in the consort flow diagram in Figure~\ref{fig:consortflow}. Participant demographics are summarized in Table~\ref{table:demographics}. All respondents are early-to-mid career physicians with the majority being female and practicing pediatrics.

\textbf{Results:} 
Clinician confidence has a mean of 3 and a standard deviation of 0.5 and the view count per video has a mean of 1.5 and a standard deviation of 0.5. We use linear regression to analyze the correlation of both clinicians’ confidence and view count with their score. No statistically significant correlation between clinician confidence and the accuracy score (\textit{P}=.07) or with view count and score (\textit{P}=.32) were found (a P-value of $\le$ .05 was considered statistically significant).

In Table~\ref{table:results_doc}, we compare anomaly detection methods against clinicians in accuracy, sensitivity, specificity and precision. We compute the metrics for all individual clinicians and then compute the mean and standard deviation of the group. We choose the decision threshold of machine learning approaches of which the sensitivity (TPR) on the validation set to be 90\%, and then report the metric values on the testing set. Compared to clinicians, our methods perform better and achieve lower variances across all metrics. We further notice that SCAD-CJ-WF outperforms both clinicians and all other methods in most metrics. 

\begin{table}[ht]
\centering
\begin{tabular}[t]{l c c c c c} 
    \toprule
    {Method} & {Accuracy} & {Sensitivity} & {Specificity} & {Precision}\\ 
    \midrule
    {Clinicians} & $67.8 \pm 17.0$ & $63.2 \pm 31.8$ & $72.4 \pm 17.4$ & $64.5 \pm 24.3$\\
    \midrule
    {MSC} & $51.7 \pm 7.6$ & $43.3 \pm 15.3$ & $60.0 \pm 0.0$ & $51.0 \pm 8.6$\\
    {SCAD-CJ-RC} & $73.3 \pm 2.9$ & $70.0 \pm 0.0$ & $76.7 \pm 5.8$ & $75.2 \pm 4.5$\\
    {SCAD-CJ-RR} & $78.3 \pm 7.6$ & $\mathbf{90.0 \pm 0.0}$ & $66.7 \pm 15.3$ & $73.7 \pm 8.8$\\
    {SCAD-CJ-WF} & $\mathbf{81.7 \pm 2.9}$ & $83.3 \pm 5.8$ & $\mathbf{80.0 \pm 0.0}$ & $\mathbf{80.6 \pm 1.0}$\\
    \bottomrule
\end{tabular}
\caption{Comparison of machine learning models against a group of clinicians (n=25) in performances (mean and standard deviation of accuracy \%, sensitivity \%, specificity \% and precision \%) on testing set. Machine learning models' decision thresholds are chosen by setting the sensitivity (TPR) on the validation set to be 90\%.}
\label{table:results_doc}
\end{table}

\begin{figure}
  \centering
    \includegraphics[width=3.25in]{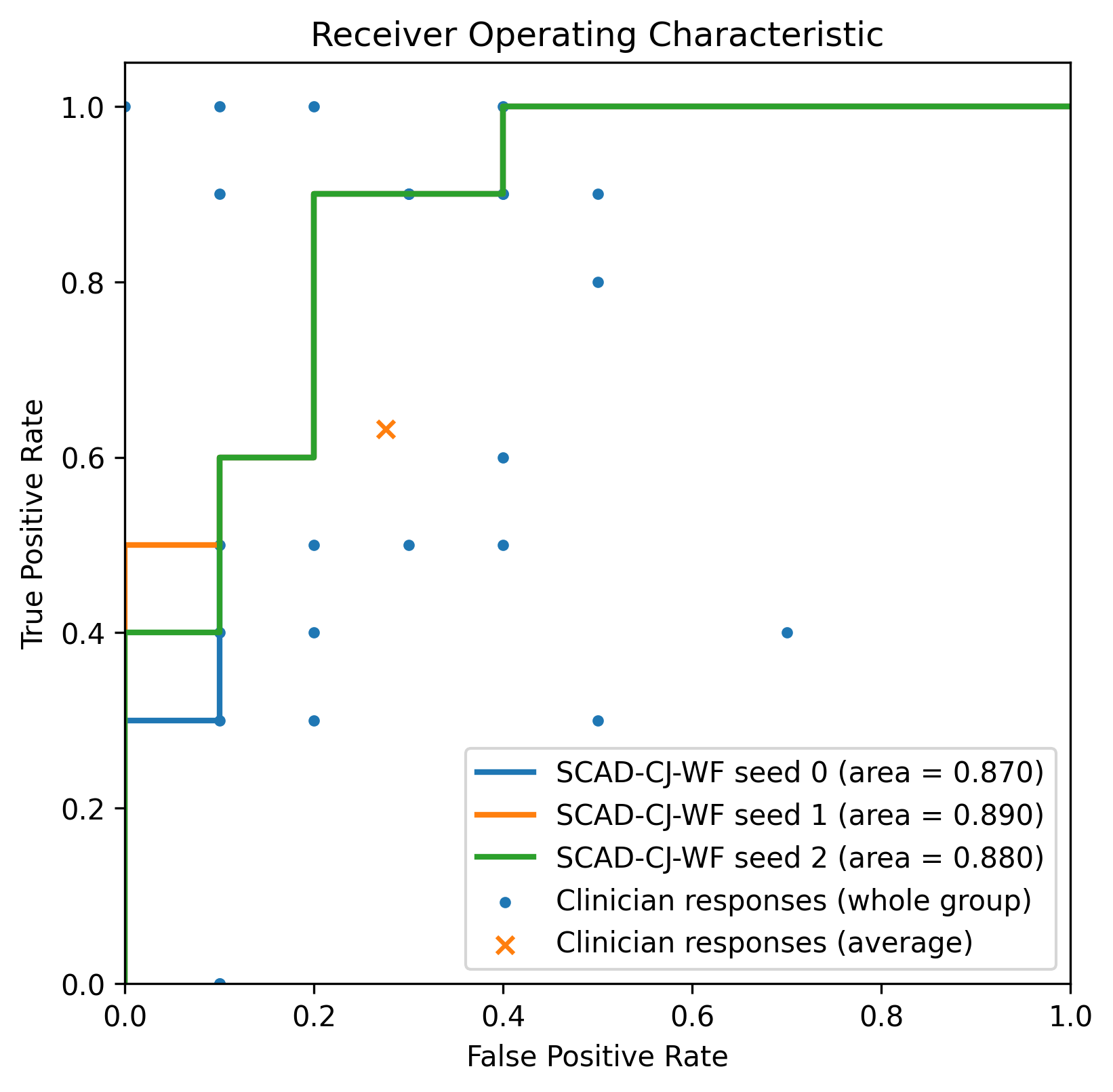}
    \caption{Overlaying clinicians' TPR/FPR samples (n=25) and group average on SCAD-CF-WF's Receiver Operative Characteristic (ROC) curve. Multiple clinicians' responses may have the same TPR/FPR values.}
  \label{fig:roc}
\end{figure}

In Figure~\ref{fig:roc}, we provide further visualizations comparing our best performing model SCAD-CJ-WF to clinicians. All seeds of our model produced an ROC curve that exceeds the average clinician response. Controlling the same False Positive Rate, our model outperforms the clinician performance in True Positive Rate (Sensitivity) by a large margin from 61.5\% to 90\%. Besides the improvement in average performance, our model also generates more consistent predictions. We can see from the figure that different clinicians produce results that are not consistent with each other. While the best clinician included in our study did achieve a perfect True and False Positive Rate, the majority of clinician evaluations are scattered with a large variation. The wide variability and lower accuracy of clinician's scores from this sample are comparable to prior reports of pediatricians and general practitioners at identifying AOM from otoscopy~\cite{pichichero2005comparison, pichichero2001assessing}. This variability between clinician scores reflects the clinical challenge of accurately diagnosing AOM. Individual factors such as a clinicians' training and experience, combined with patient factors such as cooperativeness and a non-obstructed view contribute to a successful diagnosis. In contrast, our model across different seeds generated more consistent and more accurate results. Given the clinician variability to diagnose AOM illustrated our comparative study and previous studies, there is significant value to improve diagnosis in a clinical setting by adopting deep learning based anomaly detection screening like that of the present study. 

\textbf{Discussion:} Our presented method shows promising performance in anomaly detection on otoscopy video sequences. The feasibility of adopting computer-aided diagnostics to detect and report anomaly may assist front-line clinicians in primary care and eventually even patients at-home. 

The eventual successful deployment of such technology necessitates more than the successful development of an diagnostic AI algorithm with high accuracy. For such technology to integrate within current healthcare models and work flow there is a need to create a whole new technological infrastructure. The implementation of such technology beyond it obvious need to be adopted by healthcare providers, perhaps the more pressing questions where the required financial investment for the creation of this infrastructure will come from. 

Meanwhile, our study exhibits several limitations at the current stage. i) The dataset is relatively small, skewed towards normal and collected through convenience sampling. As such our data serve to provide some additional support for the proof of concept but does not negate the need for larger and more rigorously collected data-set before such technology becomes clinically viable. The development of a large video database would likely further improve the algorithm's diagnostic performance and enable training of specific diagnosis as apposed to being limited to detecting normal vs abnormal ear exams. ii) Cellscope oto does not permit pneumotoscopy (blow air at eardrum to evaluate for movement) which can also serve as a critical step for accurate diagnosis of ear disease. iii) In our comparative study clinicians were not provided with patient history which might increase their diagnostic accuracy in a clinical setting. 

\section{Conclusion}
\IEEEPARstart{I}{n} this study, we present a two-stage anomaly detection method that is designed to perform normal / abnormal classification for otoscopy videos and that can be developed based on small dataset and highly skewed class distribution. We demonstrate that our method outperforms baseline algorithms as well as the average of 25 clinicians, constituting a promising step towards computer-aided diagnosis of the middle and external ear pathology. The further development of computer-aided diagnosis method, like ours, could contribute to timely and high-quality management of ear disease. 
\section*{Acknowledgment}
The authors gratefully acknowledge funding by the Leon Lowenstein Foundation.

\ifCLASSOPTIONcaptionsoff
  \newpage
\fi





\bibliographystyle{IEEEtran}
\bibliography{IEEEabrv,Bibliography}

\vfill

\end{document}